# Pignistic Probability Transforms
# for Mixes of Low- and High-Probability Events

John J. Sudano
Lockheed Martin
Moorestown, NJ, 08057, USA
john.j.sudano@lmco.com

**Abstract-***In some real world information fusion situations, time critical decisions must be made with an incomplete information set. Belief function theories (e.g., Dempster-Shafer theory of evidence, Transferable Belief Model) have been shown to provide a reasonable methodology for processing or fusing the quantitative clues or information measurements that form the incomplete information set.*

*For decision making, the pignistic (from the Latin pignus, a bet) probability transform has been shown to be a good method of using Beliefs or basic belief assignments (BBAs) to make decisions. For many systems, one need only address the most-probable elements in the set. For some critical systems, one must evaluate the risk of wrong decisions and establish safe probability thresholds for decision making. This adds a greater complexity to decision making, since one must address all elements in the set that are above the risk decision threshold. The problem is greatly simplified if most of the probabilities fall below this threshold. Finding a probability transform that properly represents mixes of low- and high-probability events is essential.* ***This article introduces four new pignistic probability transforms with an implementation that uses the latest values of Beliefs, Plausibilities, or BBAs to improve the pignistic probability estimates.*** *Some of them assign smaller values of probabilities for smaller values of Beliefs or BBAs than the Smets pignistic transform. They also assign higher probability values for larger values of Beliefs or BBAs than the Smets pignistic transform. These probability transforms will assign a value of probability that converges faster to the values below the risk threshold.*

*A probability information content (PIC) variable is also introduced that assigns an information content value to any set of probability. Four operators are defined to help simplify the derivations.*

***This article outlines a systematic methodology of making better decisions using Belief function theories. This methodology can be used to automate critical decisions in complex systems.***

# 1 Introduction

For some critical systems, one must evaluate the risk of wrong decisions and establish safe probability thresholds for decision making. This adds greater complexity to decision making, since one must address all elements in the set that are above the risk decision threshold. The problem is greatly simplified if most of the probabilities fall below this threshold. Finding a probability transform that properly represents mixes of low- and high-probability events is essential. This article introduces four new pignistic probability transforms. Some of them assign smaller values of probabilities for smaller values of Beliefs or BBAs than the Smets pignistic transform. They also assign higher probability values for larger values of Beliefs or BBAs than the Smets pignistic transform. These probability transforms will assign a value of probability that converges faster to the values below the risk threshold.

## 1.1 Background

Let $\Omega$ be the set of possible outcomes, where the outcomes are mutually exclusive and exhaustive singleton elements of the decision environment. In Bayesian formalism, the probabilities are assigned only to the singleton subsets of the quantitative information set. These probabilities are used to make the decisions.

For systems with a complex input (real-time sensor measurements, multidimensional filtered feature extractions, real-time data base and *a priori* data base information content, natural language text and symbols parsing evidence, quantitative and qualitative communication clues, and inconsistent errors), a Power-set $(\Omega) = 2^{\Omega}$ representation of the outcomes and a two-level (lower & upper) probability portrayal is a better representation of the incomplete information set; i.e., some sensor measurements of attributes populate more than one hypothesis.

Belief theories maintain a two-level probabilistic portrayal of the information set: the Belief or credal level and the Plausibility level. The primary foundation in any decision preposition $A_i$ is the value of the Belief Bel($A_i$), while the plausibility $Pl(A_i)$ provides the secondary support for the decision.

The basic belief assignment $m(A_J)$ represents the strength of all the incomplete information set for the outcome $A_J$. The assignments of the BBAs values to a specific subset $A_J \in 2^\Omega$ are constrained by the normalization constraint equation.

$$\sum_{A_J \in 2^\Omega} m(A_J) = 1$$

Using the BBAs as the representative of the incomplete information set, the Belief function can be computed. The Belief of the subset preposition $A_J$ is the sum $m(A_K)$ for all the subsets of $A_K$ containes in $A_J$.

$$Bel(A_J) = \sum_{A_K \supseteq A_J} m(A_K)$$

The Plausibility of the subset $A_J$ is the sum $m(A_K)$ for all subsets $A_K$ that have a non-null intersection of $A_J$.

$$Pl(A_J) = \sum_{A_K \cap A_J \neq 0} m(A_K)$$

For any singleton proposition $A_i$, the probability is bound between the Belief and the Plausibility.

$$Bel(A_i) \leq Probability(A_i) \leq Pl(A_i)$$

## 1.2 Simplifying Notation

Define the singleton subsets with a lower case letter index and define compound subsets with a capital letter index. See the following example where the set consists of three elements:

Let the set $\Omega = \{A_i, A_j, A_k\}$ with power-set

$$2^\Omega = \{f, A_i, A_j, A_k, \{A_i, A_j\}, \{A_i, A_k\}, \{A_j, A_k\}, \{A_i, A_j, A_k\}\}$$

$$A_L = \{A_i, A_j\}$$
$$A_M = \{A_i, A_k\},$$
$$A_N = \{A_j, A_k\},$$
$$A_P = \{A_i, A_j, A_k\}$$

With the above definitions the power-set is $2^\Omega = \{f, A_i, A_j, A_k, A_L, A_M, A_N, A_P\}$

In order to simplify the derivations, four operators are introduced: the union-to-sum operator, the sum-to-union operator, the compound-to-sum of singletons operator, and the sum-to-compound operator. These operators expand and contract the operations on elements of the set.

The union-to-sum operator, $\vec{U}^S$, on singleton elements has the following properties:

$$\vec{U}^S[Bel(A_j \cup A_k \cup ... \cup A_z)] \equiv Bel(A_j) + Bel(A_k) + ... + Bel(A_z)$$
$$\vec{U}^S[Pl(A_j \cup A_k \cup ... \cup A_z)] \equiv Pl(A_j) + Pl(A_k) + ... + Pl(A_z)$$
$$\vec{U}^S[m(A_j \cup A_k \cup ... \cup A_z)] \equiv m(A_j) + m(A_k) + ... + m(A_z)$$

The sum-to-union operator, $\vec{S}^U$, on singleton elements has the following properties:

$$\vec{S}^U[Bel(A_j) + Bel(A_k) + ... + Bel(A_z)] \equiv Bel(A_j \cup A_k \cup ... \cup A_z)$$
$$\vec{S}^U[Pl(A_j) + Pl(A_k) + ... + Pl(A_z)] \equiv Pl(A_j \cup A_k \cup ... \cup A_z)$$
$$\vec{S}^U[m(A_j) + m(A_k) + ... + m(A_z)] \equiv m(A_j \cup A_k \cup ... \cup A_z)$$

The compound-to-sum operator, $\vec{C}^S$, on singleton elements has the following properties:

$$\vec{C}^S[Bel(\{A_j, A_k, ..., A_z\})] \equiv Bel(A_j) + Bel(A_k) + ... + Bel(A_z)$$
$$\vec{C}^S[Pl(\{A_j, A_k, ..., A_z\})] \equiv Pl(A_j) + Pl(A_k) + ... + Pl(A_z)$$
$$\vec{C}^S[m(\{A_j, A_k, ..., A_z\})] \equiv m(A_j) + m(A_k) + ... + m(A_z)$$

The sum-to-compound operator, $\vec{S}^C$, on singleton elements has the following properties:

$$\vec{S}^C[Bel(A_J) + Bel(A_K) + ... + Bel(A_Z)] \equiv Bel(\{A_J, A_K, ..., A_Z\})$$
$$\vec{S}^C[Pl(A_j) + Pl(A_k) + ... + Pl(A_z)] \equiv Pl(\{A_j, A_k, ..., A_z\})$$
$$\vec{S}^C[m(A_j) + m(A_k) + ... + m(A_z)] \equiv m(\{A_j, A_k, ..., A_z\})$$

## 2 Probability Information Content

A metric depicting the strength of a critical decision by a specific probability distribution is an esential measure in any threshold-driven automated decision system. A value of one would indicate the total knowledge or information to make a correct decision (one hypothesis would have a probability value of one and the rest of zero). A value of zero would indicate that the knowledge or information to make a correct decision does not exist (all the hypothesis would have an equal probability value).

The Kullback-Leibler divergence is a measure of the difference between two probability distributions P and Q.

$$D(P,Q) = \sum_{i=1}^{N} p(i) Log(p(i)/q(i))$$

The uniform distribution, U, depicts a system that has all the hypotheses with an equal probability value so that a correct decision can not be made (it conveys no information).

$$\sum_{i=1}^{N} u(i) = 1 \quad \text{with} \quad u(i) = \frac{1}{N}$$

On comparing any distribution to the uniform distribution

$$D(P,U) = \sum_{i=1}^{N} p(i) \ Log \ p(i) - \sum_{i=1}^{N} p(i) \ Log \ u(i)$$

$$D(P,U) = \sum_{i=1}^{N} p(i) \ Log \ p(i) + Log \ N$$

Normalizing the above equation allows the comparison of the information content of any distribution to that of the uniform distribution. Therefore, a **probability information content (PIC) variable is introduced** that assigns a value to any set of probabilities. If there are N possible hypothesis $\{H_1, H_2, ..., H_N\}$ with respective probabilities $\{P(1), P(2), ..., P(N)\}$, then the probability information content is:

$$PIC \equiv 1 + \frac{\sum_{i=1}^{N} P(i) Log[P(i)]}{Log[N]}$$

A PIC value of one indicates total knowledge is available and there is no ambiguity in the decision making; i.e., one of the hypotheses has a probability value of one and the rest have zero. A PIC value of zero indicates that all the hypotheses have an equal probability of occurring and it is not possible to make a good decision with the probability distribution that is available.

Example: An encrypted data stream should have a PIC value of zero, while a decrypted data stream should have a PIC value of one.

# 3 The Rationale of Mapping Basic Belief Assignments to Probabilities

Smets gives a rationale to compute a pignistic probability transform by using Johann Bernoulli's insufficient-reason principle: *If one is ignorant of the ways an event can occur and therefore has no reason to believe that one way will occur preferentially to another, it will occur equally likely in any way.* Hence, the BBAs are distributed equally among each contributing probabilistic element.

Since the basic belief assignments, Beliefs, and Plausibilities exist at the time of decision, this information can be used to aid the process of computing the probability. This article introduces four new pignistic probability transforms that use the knowledge of the basic belief assignments, Beliefs, and/or Plausibilities {See (3.2),(3.3),(3.4),(3.5)} to distribute the BBAs proportional to the prior knowledge among each contributing probability element.

For each singleton element $A_i$, the probability is bound between the Belief and the Plausibility.

$$\text{Bel}(A_i) \leq \text{Probability}(A_i) \leq \text{Pl}(A_i)$$

The probability is normalized to one.

$$\sum_{A_i \subseteq \Omega} \text{Probability}(A_i) = 1$$

## 3.1 The Smets Pignistic Probability (BetP)

The Smets pignistic probability (SPP) transformation uses the BBAs to assign a pignistic probability to the singleton subsets. The pignistic transformation hypothesis builds a probability distribution on n elements and given a lack of information, gives a probability 1/n to each element. This procedure is repeated for each basic belief assignment $m(\{A_1, A_2, A_3, ..., A_n\})$ and is distributed equally among each element of $A_i$ for all $A_i \subseteq \Omega$. Therefore, Smets pignistic transformation hypothesis distributes the basic belief assignments $m(A_J)$ equally among each singleton element of $A_i \subseteq A_J$, with $A_i \subseteq \Omega$ for all $A_J \in 2^\Omega$.

The Smets pignistic probability equation is:

$$BetP(A_i) = \sum_{A_M \supseteq A_i} \left(\frac{1}{|A_M|}\right) m(A_M)$$

$$i = 1, 2, ..., N$$

where $|A_M|$ is the number of singleton elements in $A_M$. Note that SPP is normalized to one.

$$\sum_{A_i \subseteq \Omega} BetP(A_i) = 1$$

For each singleton element $A_i$, the pignistic probability is bound between the Belief and the Plausibility.

$$\text{Bel}(A_i) \leq \text{BetP}(A_i) \leq \text{Pl}(A_i)$$

## 3.2 Probability Deficiency is Proportional to all Plausibilities (PraPl)

The pignistic probability can be approximated by the Belief and a component proportional to the sum of all the Plausibilities (PraPl).

$$\text{PraPl}(A_i) \equiv \text{Bel}(A_i) + e\ \text{Pl}(A_i)$$

with $\quad e = \dfrac{1 - \sum_{A_i \subseteq \Omega} \text{Bel}(A_i)}{\sum_{A_i \subseteq \Omega} \text{Pl}(A_i)}$

This transformation distributes the basic belief assignments $m(A_J)$ equally among each singleton probability element. The pignistic probability is normalized to one.

$$\sum_{A_i \subseteq \Omega} \text{PraPl}(A_i) = 1$$

For each singleton element $A_i$, the pignistic probability is bound between the Belief and the Plausibility.

$$\text{Bel}(A_i) \leq \text{PraP}(A_i) \leq \text{Pl}(A_i)$$

## 3.3 Proportional Plausibility Probability (PrPl)

The proportional Plausibility (PrPl) transformation hypothesis assumes that the basic belief assignments $m(A_J)$ are distributed proportionally to the plausibility among each singleton element of $A_i \subseteq A_J$ with $A_i \subseteq \Omega$ for all $A_J \in 2^\Omega$.

$$\text{PrPl}(A_i) = \sum_{A_M \supseteq A_i} \left( \frac{Pl(A_i)}{\vec{C}^S[Pl(A_M)]} \right) m(A_M)$$

The pignistic probability is normalized to one.

$$\sum_{A_i \subseteq \Omega} \text{PrPl}(A_i) = 1$$

For each singleton element $A_i$, the pignistic probability is bound between the Belief and the Plausibility.

$$\text{Bel}(A_i) \leq \text{PrPl}(A_i) \leq \text{Pl}(A_i)$$

## 3.4 Proportional Belief Probability (PrBl)

The proportional Belief (PrBl) transformation hypothesis assumes that the basic belief assignments $m(A_J)$ are distributed proportionally to the Belief among each singleton element of $A_i \subseteq A_J$ with $A_i \subseteq \Omega$ for all $A_J \in 2^\Omega$. Note that for a singleton element, $\text{Bel}(A_i) = m(A_i)$.

$$\text{PrBl}(A_i) = \sum_{A_M \supseteq A_i} \left( \frac{Bel(A_i)}{\vec{C}^S[Bel(A_M)]} \right) m(A_M)$$

$$\text{PrBl}(A_i) = \sum_{A_M \supseteq A_i} \left( \frac{m(A_i)}{\vec{C}^S[m(A_M)]} \right) m(A_M)$$

The pignistic probability is normalized to one.

$$\sum_{A_i \subseteq \Omega} \text{PrBl}(A_i) = 1$$

For each singleton element $A_i$, the pignistic probability is bound between the Belief and the Plausibility.

$$\text{Bel}(A_i) \leq \text{PrBl}(A_i) \leq \text{Pl}(A_i)$$

## 3.5 Proportional Self-Consistent Probability

The proportional to the self consistent pignistic probability (PrScP) transformation hypothesis, introduced in this article, assumes that the BBAs are distributed proportionally to the probability among each singleton element of $A_i \subseteq \Omega$ and $A_M \in 2^\Omega$.

$$\text{PrScP}(A_i) = \sum_{A_M \supseteq A_i} \left( \frac{\text{PrScP}(A_i)}{\vec{C}^S[\text{PrScP}(A_M)]} \right) m(A_M)$$

Note that the solution to the above equation is contained on both sides of the equation. This equation must be solved numerically. A simple method of a solution is the self-consistent method: (1) Initiate the PrScPs with PrBl; (2) Calculate the new PrScP using the above equation and old values of PrScP; (3) repeat step two until convergence.

The self-consistent pignistic probability is normalized.

$$\sum_{A_i \subseteq \Omega} \text{PrScP}(A_i) = 1$$

For each singleton element $A_i$, the pignistic probability is bound between the Belief and the Plausibility.

$$\text{Bel}(A_i) \leq \text{PrScP}(A_i) \leq \text{Pl}(A_i)$$

## 4 A Discussion of $m(\{A_1, A_2\})$

$m(A_1, A_2)$ provides support to the singleton probability hypothesis $A_1$ and $A_2$. The generalized breakdown to the contributing $A_1$ and $A_2$ probabilities are of the form:

$$\text{Probability}(A_1) \approx a \ m(\{A_1, A_2\})$$
$$\text{Probability}(A_2) \approx (1-a) \ m(\{A_1, A_2\})$$

Realistically $a$ can have a value between zero and one ($0 \leq a \leq 1$), depending on the type of information used to populate $m(\{A_1, A_2\})$. If the total information used is quantitative, then a correct value of $a$ can be computed. However, if the real world incomplete information set is a complex mix of qualitative, quantitative, conflicting, and deceptive data, then the value of $a$ cannot be computed analytically, but must be estimated using conjectured rationale. The results must be interpreted accordingly.

For the Smets pignistic probability: $a = \dfrac{1}{2}$

For the proportional Plausibility probability (PrPl):

$$a = \frac{Pl(A_1)}{Pl(A_1) + Pl(A_2)}$$

For the proportional Belief probability (PrBl):

$$a = \frac{m(A_1)}{m(A_1) + m(A_2)} = \frac{Bel(A_1)}{Bel(A_1) + Bel(A_2)}$$

For the self-consistent pignistic probability (PrScP):

$$a = \frac{\text{PrScP}(A_1)}{\text{PrScP}(A_1) + \text{PrScP}(A_2)}$$

For the Probability Belief difference proportional to the sum of all Plausibilities, the contribution of $m(\{A_1, A_2\})$ provides an equal contribution to both probabilities via the plausibility values of $Pl(A_1)$ and $Pl(A_2)$ as described in the following equations:

$$\text{PraPl}(A_1) = \text{Bel}(A_1) + e \ \text{Pl}(A_1)$$
$$\text{PraPl}(A_2) = \text{Bel}(A_2) + e \ \text{Pl}(A_2)$$

with $\quad e = \dfrac{1 - \sum_{A_i \subseteq \Omega} \text{Bel}(A_i)}{\sum_{A_i \subseteq \Omega} \text{Pl}(A_i)}$

## 5 Example: Combat Identification

Combat Identification (Friend, Neutral, Hostile, and Unknown) is a very complex problem with high-risk consequences for erroneous decisions. For a given identification (ID) of Friend, Neutral or Hostile, the response is normally a preprogrammed response. For an identification of Unknown, the response is laden with a high-risk consequence.

The following example illustrates some design concepts presented in this article. In a certain geographical location, a multi-source integration (MSI) using origin & flight evidence, quantitative sensor measurement, and feature derived estimates that form the incomplete information set taxonomy, the identification BBAs representation is given as:

$m(F) = 0.16 \qquad m(N) = 0.14$
$m(U) = 0.01 \qquad m(H) = 0.02$
$m(\{F,N\}) = 0.20 \qquad m(\{F,U\}) = 0.09$
$m(\{F,H\}) = 0.04 \qquad m(\{N,U\}) = 0.04$
$m(\{N,H\}) = 0.02 \qquad m(\{U,H\}) = 0.01$
$m(\{F,N,U\}) = 0.10$
$m(\{F,N,H\}) = 0.03$
$m(\{F,U,H\}) = 0.03$
$m(\{N,U,H\}) = 0.03$
$m(\{F,N,U,H\}) = 0.08$

For these BBAs the Beliefs are calculated to be:

$Bel(F) = 0.16 \qquad Bel(N) = 0.14$

$Bel(U) = 0.01 \qquad Bel(H) = 0.02$

Having a Belief sum of **0.33.**

For these BBAs the Plausibilities are calculated to be:

$Pl(F) = 0.73 \qquad Pl(N) = 0.64$

$Pl(U) = 0.39 \qquad Pl(H) = 0.26$

Having a Plausibility sum of **2.02.**

At decision time, the Smets pignistic probabilities are computed from the above BBAs as:

$BetP\ (F) = 0.398333$
$BetP\ (N) = 0.343333$
$BetP\ (U) = 0.153333$
$BetP\ (H) = 0.105000$

The above pignistic probability has a probability information content (PIC) value of **0.092643**

At decision time, the pignistic probability that is assumed to be proportional to the sum of all the Plausibilities (PraPl) is computed as:

$PraPl\ (F) = 0.402129$
$PraPl\ (N) = 0.352277$
$PraPl\ (U) = 0.139356$
$PraPl\ (H) = 0.106238$

with $\quad a = 0.331683$

The above probability set has a probability information content value of **0.100695**

The proportional to Plausibility probabilities (PrPl) are also computed as:

$PrPl\ (F) = 0.454418$
$PrPl\ (N) = 0.360880$
$PrPl\ (U) = 0.117638$
$PrPl\ (H) = 0.067064$

The above probability set has a probability information content value of **0.163811**

The proportional Belief probabilities (PrBl) are also computed as:

$PrBl\ (F) = 0.517592$
$PrBl\ (N) = 0.405098$
$PrBl\ (U) = 0.030288$
$PrBl\ (H) = 0.047022$

The above probability set has a probability information content value of **0.309962**

The proportional self-consistent pignistic probabilities (PrScP) are computed self-consistently as:

$PrScP\ (F) = 0.542030$
$PrScP\ (N) = 0.386953$
$PrScP\ (U) = 0.032397$
$PrScP\ (H) = 0.038620$

The above probability set has a probability information content value of **0.324722**

A comparison of these pignistic probabilities demonstrate the differences, for an *a priori* decision threshold of **0.0455**:

The Smets pignistic probability of all four states are higher than the threshold and all four IDs are displayed to an human operator.

The pignistic probability assumed to be proportional to the sum of all the Plausibilities (PraPl) has all four probability values higher than the threshold, therefore all four IDs are displayed to an human operator.

The pignistic probability assumed to be proportional to the the Plausibility (PrPl) has also all four probabilities higher than the threshold, therefore all four IDs are displayed to an human operator.

The pignistic probability assumed to be proportional to the the Belief (PrBl) has **three** probabilities higher than the threshold, therefore **three** IDs are displayed to an human operator.

The self-consistent pignistic probability (PrScP) of **two** states are higher than the threshold therefore, **two** IDs are displayed to an human operator.

Upon review of the MSI taxonomy, the highest supported hypothesis is that of an F-16. This platform is consistent with the information set since, in the operational geographical location, a neutral nation has 24 F-16's.

# 6 An Implementation

Picking a pignistic probability for making decisions can be acomplished by a threshold set of Beliefs and plausabilities functions. Elements in this threshold set are computed by sums, maximum, and minimum of Beliefs & Plausabilities. The set of elements with appropriate logic are used to choose the pignistic probability.

A simple implementation is to use the values of the sum of the Beliefs and Plausibilities at the time of decision to evaluate the maturity of the incomplete information set. These values would dictate which pignistic transformation that should be used.

$$SumBel = \sum_{A_i \subseteq \Omega} Bel(A_i)$$
$$SumPl = \sum_{A_i \subseteq \Omega} Pl(A_i)$$

Let a set of Belief thresholds be: $\{ Bel_1^T < Bel_2^T < Bel_3^T \}$ and the Plausibility thresholds be: $\{ Pl_1^T < Pl_2^T < Pl_3^T \}$ then the following logic is used in an implementation.

If( $SumBel > Bel_3^T$ & $SumPl < Pl_1^T$ ) **Then** use the self-consistent pignistic probability transform (PrScP)

If( $SumBel > Bel_2^T$ & $SumPl < Pl_2^T$ ) **Then** use the proportional to Belief pignistic probability transform (PrBl).

If( $SumBel > Bel_1^T$ & $SumPl < Pl_3^T$ ) **Then** use the proportional to Plausibilities pignistic probability transform (PrPl).

**Else** use Smets pignistic probability transform.

For any application there is more than one set of thresholds depending on the system situation: (1) an automobile may have one set for highway driving, one for off road, and another for city driving; (2) A missile may have a set for each type class of threat; (3) A market analysis program may have a set for a bull market, a set for a transitional market, and another for a bear market; (4) A combat identification system may have a set for each level of operation: peaceful, imminent conflict, and war; (4) A tumor diagnostic system may have a set for each type of expected class of tumors.

## 7 Conclusion:

In some real world information fusion situations, time critical decisions must be made with an incomplete information set. Belief function theories (e.g., Dempster-Shafer theory of evidence, Transferable Belief Model) have been shown to provide a reasonable methodology for processing or fusing the quantitative clues or information measurements that form the incomplete information set.

For decision making, a pignistic probability transform has been shown to be a good method of using Beliefs or BBAs to make decisions. For some critical systems, one must evaluate the risk of wrong decisions and establish safe probability thresholds for decision making. This adds a greater complexity to decision making, since one must address all elements in the set that are above the risk decision threshold.

Four new operators have been introduced to simplify derivations, the union-to-sum operator, the sum-to-union operator, compound-to-sum operator, and the sum-to-compound operator.

This article introduced four new pignistic probability transforms that may be used to make decisions depending on the maturity of the incomplete information set. The probability information content (PIC) variable was also introduced in this article that can be used to evaluate the various pignistic probability transforms. Some of these transforms assign smaller values of probabilities for smaller values of Beliefs or BBAs than Smets pignistic transform. They also assign higher probability values for larger values of Beliefs or BBAs than the Smets pignistic transform. These probability transforms will assign a value of probability that converges faster to the values below the risk threshold. A rationale for their construct was also given.

The PIC variable has been shown to have the correct limits. A PIC value of one indicates total knowledge is available and there is no ambiguity in the decision making. (i.e., one of the hypotheses has a probability value of one and the rest have zero.) A PIC value of zero indicates that all the hypothesis have an equal probability of occurring and it is impossible to make a good decision with the probability distribution that is available.

**This article outlines a systematic methodology of making better decisions using Belief function theories. This methodology can be used to automate critical decisions in complex systems.**

## References:

E.A. Bender, "Mathematical Methods in Artificial Intelligence," IEEE Computer Society Press (1996)

F. Gambino, G. Ulivi, M. Vendittelli, "The Transferable Belief Model in Ultrasonic Map Building," 1997, Proceedings of the Sixth IEEE Conference on Fuzzy Systems, vol. 1, pages 601-608, July 1-5, 1997

M. Haft, R. Hofmann, V. Tresp, "Model-Independent Mean Field Theory as a Local Method for Approximate Propagation of Information," Computation in Neural Systems, pages 93-105, 1999.

S. Kullback, *Information Theory and Statics,* Wiley, New York (1959)

O. Moseler, D. Juricic, A. Rakar, N. Muller, "Model-Based Fault Diagnosis of an Actuator System Driven by the Brushless DC Motor," 1999 Proceedings of the American Control Conference, vol. 6, pages 3779-3783, June 2-4, 1999.

P. Smets, R. Kennes, "The Transferable Belief Model," Artificial Intelligence, vol. 66, pages 191-243, 1994.